\documentclass[runningheads]{llncs}
\usepackage[dvipdfmx]{graphicx}
\usepackage[dvipdfmx]{color}

\usepackage{tikz}
\usepackage{comment}
\usepackage{amsmath} 
\usepackage{amssymb} 
\usepackage[subrefformat=parens]{subcaption}

\usepackage{amsmath,graphicx}
\usepackage{booktabs}

\usepackage{multirow}
\usepackage{tabto}
\usepackage{float}
\usepackage[colorlinks,
            linkcolor=blue,       
            anchorcolor=blue,  
            citecolor=blue,    
            ]{hyperref}

\usepackage{caption} 
\usepackage{array}
\usepackage{xcolor}
\usepackage{color, colortbl}
	
\definecolor{Gray}{gray}{0.9}
\definecolor{LightCyan}{rgb}{0.88,0.95,1}

\makeatletter
\newcommand*{\@rowstyle}{}
\newcommand*{\rowstyle}[1]{
  \gdef\@rowstyle{#1}%
  \@rowstyle\ignorespaces%
}
\newcolumntype{=}{
  >{\gdef\@rowstyle{}}%
}
\newcolumntype{+}{
  >{\@rowstyle}%
}
\makeatother

\usepackage{amssymb}
\usepackage{pifont}

\begin{document}

\title{
Bridge Damage Cause Estimation Using Multiple Images Based on Visual Question Answering}
\titlerunning{Bridge Damage Cause Estimation Using Multiple Images Based on VQA}

\author{
Tatsuro Yamane\inst{1} 
\and
Pang-jo Chun\inst{2} 
\and
Ji Dang\inst{3} 
\and
Takayuki Okatani\inst{4,5} 
}

\authorrunning{Tatsuro Yamane et al.}

\institute{Graduate School of Frontier Sciences, The University of Tokyo\and
Graduate School of Engineering, The University of Tokyo
\and
Graduate School of Science and Engineering, Saitama University
\and
Graduate School of Information Sciences, Tohoku University
\and
RIKEN Center for AIP \\
\email{yamane.tatsuro.20@dois.k.u-tokyo.ac.jp}}

\maketitle


\begin{abstract}
In this paper, a bridge member damage cause estimation framework is proposed by calculating the image position using Structure from Motion (SfM) and acquiring its information via Visual Question Answering (VQA). For this, a VQA model was developed that uses bridge images for dataset creation and outputs the damage or member name and its existence based on the images and questions. In the developed model, the correct answer rate for questions requiring the member's name and the damage's name were 67.4\% and 68.9\%, respectively. The correct answer rate for questions requiring a yes/no answer was 99.1\%. Based on the developed model, a damage cause estimation method was proposed.  In the proposed method, the damage causes are narrowed down by inputting new questions to the VQA model, which are determined based on the surrounding images obtained via SfM and the results of the VQA model. Subsequently, the proposed method was then applied to an actual bridge and shown to be capable of determining damage and estimating its cause. The proposed method could be used to prevent damage causes from being overlooked, and practitioners could determine inspection focus areas, which could contribute to the improvement of maintenance techniques. In the future, it is expected to contribute to infrastructure diagnosis automation.
\end{abstract}


\section{Introduction}
Aging infrastructure such as bridges and tunnels can cause serious accidents if left unattended. According to the ASCE's 2021 Infrastructure Report Card~\cite{asce2021infrastructure}, there are more than 617,000 bridges in the United States, 42\% of which are more than 50 years old, and 7.5\%, or 46,154 bridges, are structurally deficient or in `poor' condition. Owning to this pressing issue of aging infrastructure, there has been increased research interest in proper infrastructure safety assessment~\cite{ou2010structural,qarib2014recent}. In addition, regular condition assessment and monitoring are important to properly assess infrastructure safety~\cite{amezquita2015feature,vaghefi2012evaluation}. However, currently, infrastructure condition assessment is performed visually, which requires extensive effort and cost~\cite{agdas2016comparison,gattulli2005condition}.

Various studies have been conducted to address this problem and realize efficient maintenance management. In recent years, research on damage detection from images using machine learning techniques has been conducted. Cha et al.~\cite{cha2017deep} proposed a crack detection method from images using convolutional neural networks (CNNs), a type of deep learning algorithm. Yang et al.~\cite{yang2018automatic} conducted a study on pixel-level crack detection using fully convolutional networks. Chun et al.~\cite{chun2021automatic1} developed a pixel-level crack detection method for concrete walls using a Light Gradient-Boosting Machine. Wang \& Hu~\cite{wang2017grid}, Zhang et al.~\cite{zhang2017automated}, and Chun et al.~\cite{chun2021automatic2} proposed a deep-learning-based crack detection method for pavements. Xu et al.~\cite{xu2018identification} proposed a damage detection framework based on restricted Boltzmann machines to detect cracks in box girders. Besides crack and fissure detection studies, various studies have been conducted on other types of damage detection. Liang~\cite{liang2019image} conducted pixel-level concrete delamination detection research. Fondevik et al.~\cite{fondevik2020image} and Rahman et al.~\cite{rahman2021semantic} proposed a pixel-level corrosion damage detection method from images.

Research is also actively being conducted to record the detected damage in a three-dimensional (3D) bridge model. Liu et al.~\cite{liu2020image} proposed a method to record cracks detected from images in a 3D pier model. In addition, Yamane et al.~\cite{yamane2022detecting} proposed a damage location recording method using an object detection model in a 3D bridge model. Other studies have been conducted on methods to record damage and non-damage areas detected from images in a 3D bridge model~\cite{yamane2023recording}.

As mentioned, many studies have been conducted on damage detection from images. Damage detection from images using the aforementioned methods is expected to reduce labour during inspections and contribute to a more efficient infrastructure maintenance. These studies contribute to damage detection automation; however, little attention is paid to how to utilize the information obtained from such images. Hence, it would be of great help to infrastructure maintenance engineers if machines could not only detect damage from images, but also estimate its cause. As an example of estimating the cause of damage, one can imagine a case where corrosion is detected, and it is estimated that it is caused by a leak from a nearby drainage pipe. As mentioned above, if the cause of damage can be estimated, it is believed that not only automated inspection, but also automated diagnosis and labour savings will be possible. Furthermore, damage cause estimation automation is useful in and of itself. For example, for inexperienced engineers, this may lead to careful confirmation of the damage cause presented to the machine, effectively preventing overlooking it. Moreover, based on suggestions from the machine, it would also contribute to improving the maintenance skills of novice engineers, as they could learn where to focus in each case.

For damage cause estimation from images, it is necessary to obtain information not only on the damage but also on related members. If a simple image classification model is used to obtain information, it is possible that the combined information in the image may be mistaken. For example, in the case of an image of the main girder and slab of a steel girder bridge, if the main girder is corroded and the slab is cracked, the model that determines the member's name may judge that it is the main girder, while the model that determines the damage name may judge that it is cracked. In this case, the individual classification model would be correct; however, it would not correctly determine which damage is occurring in which member.

An approach to obtain output that takes into account the relationship between member and damage names has been studied using an image captioning model~\cite{chun2022deep}. In that study, a model is developed to output sentences regarding the damage and the member in which the damage occurs from bridge images, making it possible to obtain information that includes the relationships between words. However, there are many possible patterns in the text that can be output based on an image. Thus, for example, if multiple damages or members are present in an image simultaneously, there is a possibility that something other than the desired information will be output.

To address this problem and obtain information from images as needed, it is necessary to use a model that can input images and questions into the model and produce output in accordance with the input content. This task is called Visual Question Answering (VQA) and has been actively studied in the field of computer vision. Meanwhile, to estimate the cause of the damage in the image, it is also necessary to acquire the surrounding conditions. However, the damage and the information related to its cause are not necessarily captured in a single image. For example, it is quite possible that an image of a particular corrosion case does not show a leaky drainage pipe that could be the cause of the damage. Therefore, when estimating the cause, it is necessary to analyse not only the image of interest but also multiple images in the surrounding area and make a decision based on the results.

Accordingly, we propose a damage cause estimation method by calculating the camera coordinates using Structure from Motion (SfM), a technique to calculate the coordinates at which images were taken, and extracting information from the images via VQA using multiple images taken around the damage based on the calculated camera coordinates. The remainder of this paper is organized as follows. First, the acquisition method for images around the image of interest using SfM is explained. Then, the training process of the VQA model using bridge images is explained. Afterward, a damage cause estimation method based on the output of the developed VQA model is explained. Field test results are presented next after applying the proposed method to an actual bridge structure. Finally, the conclusions of this study are summarized.
\section{Methodology framework of this paper}
\begin{figure}[h]
\begin{center}
\includegraphics[width=6cm]{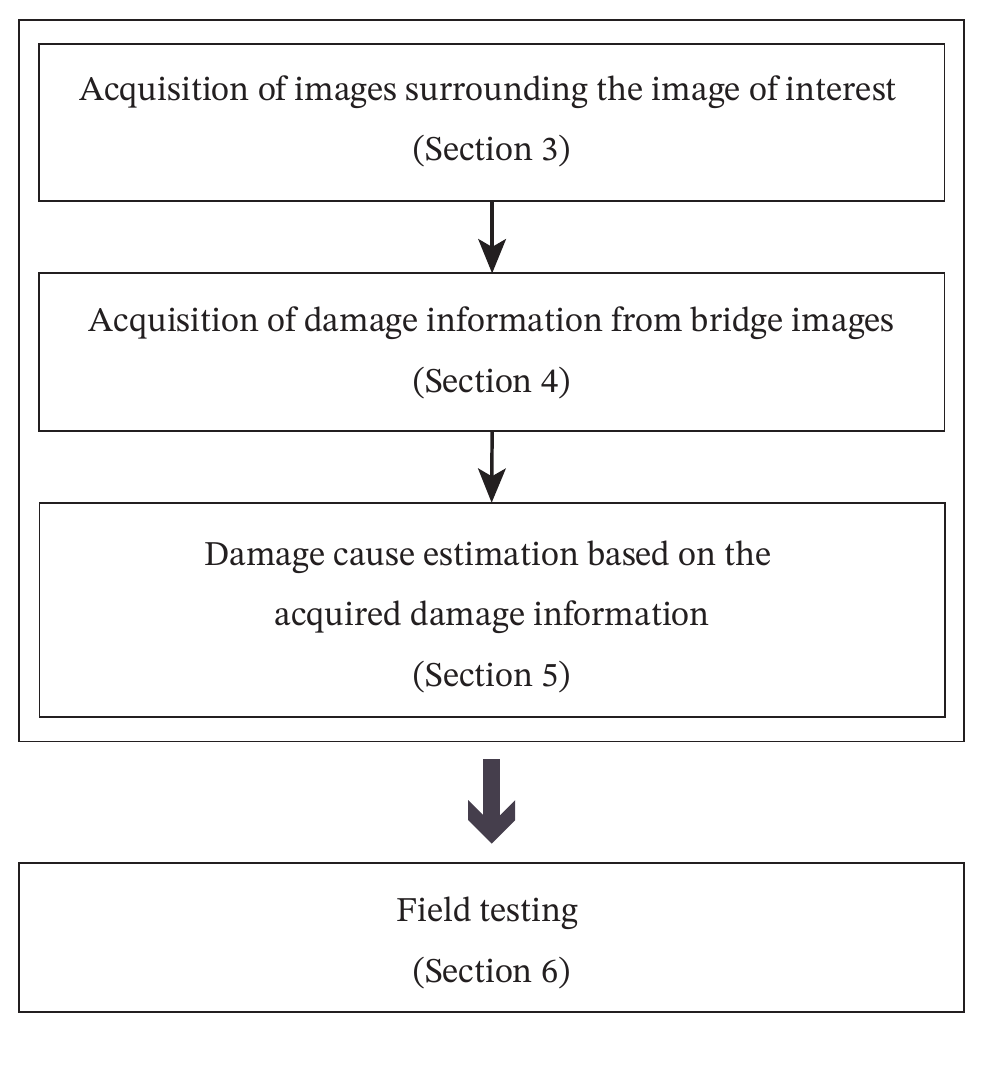}
\end{center}
   \caption{Proposed framework methodology and field test
   }
\label{fig:framework}
\end{figure}

The proposed method consists of four parts: acquisition of images surrounding the image of interest (Sec.~\ref{sec:acquisition of images}), acquisition of damage information from bridge images (Sec.~\ref{sec:acquisition of damage information}), damage cause estimation based on the acquired damage information (Sec.~\ref{sec:estimation}), and field testing (Sec.~\ref{sec:field test}). The framework is presented in Fig.~\ref{fig:framework}.

Specifically, in the acquisition of images surrounding the image of interest, we explain how to obtain the area of the bridge captured by each image based on the SfM-estimated camera position coordinates. In the acquisition of damage information from bridge images, we explain how to create a dataset for VQA using bridge images, develop a VQA model based on the created dataset, and verify the model accuracy. For damage cause estimation based on the acquired damage information, we explain how to estimate the cause of damage using the developed VQA model. Finally, we describe the field test results and discuss the application of the proposed method to an actual bridge. Each of these topics is described in detail in the following sections.

\section{Acquisition of images surrounding the image of interest}
\label{sec:acquisition of images}

\subsection{Three-dimensional scene and surface reconstruction}
In damage cause estimation, it is desirable not to use a single image alone; surrounding images are also needed. For this, the coordinates of the camera that captured each image must be calculated to determine the area of the bridge from which each image was taken. SfM is a method that can calculate the shooting coordinates of each image based on multiple images. This is a technique that detects common feature points between images from each image and calculates the camera position based on epipolar geometry. SfM is a well-known method for image-based 3D scene and surface reconstruction, a key challenge in the field of computer vision~\cite{hartley2003multiple}. The dense point cloud data of an object can be constructed based on the positions of each camera determined via SfM and stereo matching results using multiple images. The dense point cloud data can also be used to construct a polygon mesh model with a triangular mesh.

In this study, SfM is used to calculate the shooting coordinates of each image and to determine the images surrounding the image of interest. SfM is a commonly used technology, and there are various commercially available software packages for SfM-based 3D modelling. Nevertheless, the proposed framework is applicable regardless of the type of SfM software.

\subsection{Shooting area and surrounding images}
In selecting images surrounding the image of interest, it is not appropriate to simply use images whose camera coordinates are close to those of the camera that took the image of interest. This is because if the cameras are facing in different directions, it is possible that they are capturing images of areas that are far from the damage. Even if the camera coordinates are far, it is possible that the camera is capturing a close area from a different direction. Therefore, in this study, we calculate the intersection of the bridge mesh model created based on the SfM results and the camera direction vectors; this intersection is used as the coordinates of the point where the image was taken. Based on these coordinates, the surrounding images are selected.

\begin{figure}[h]
\begin{center}
\includegraphics[width=6cm]{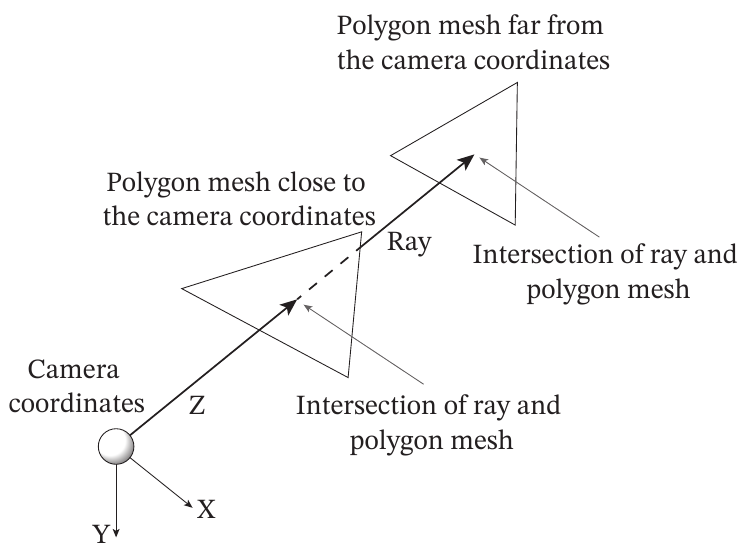}
\end{center}
   \caption{Intersection of the camera direction vector and the triangular mesh
   }
\label{fig:intersection}
\end{figure}

The mesh model created based on the SfM results consists of a set of triangular meshes. Therefore, the mesh model intersection points can be calculated by computing the triangular meshes that intersect the direction vectors of the camera whose coordinates have already been identified, as shown in Fig.~\ref{fig:intersection}. Note that the camera direction vector and the triangular mesh may intersect multiple times. Therefore, the intersection point closest to the camera coordinates is used as the shooting point coordinates.

Algorithms for determining the intersection between a triangular mesh and a ray represent a simple geometric problem and are not discussed here. For example, the M{\"o}ller Trumbore intersection algorithm~\cite{moller2005fast} can be used for intersection determination.

\begin{figure}[h]
\begin{center}
\includegraphics[width=6cm]{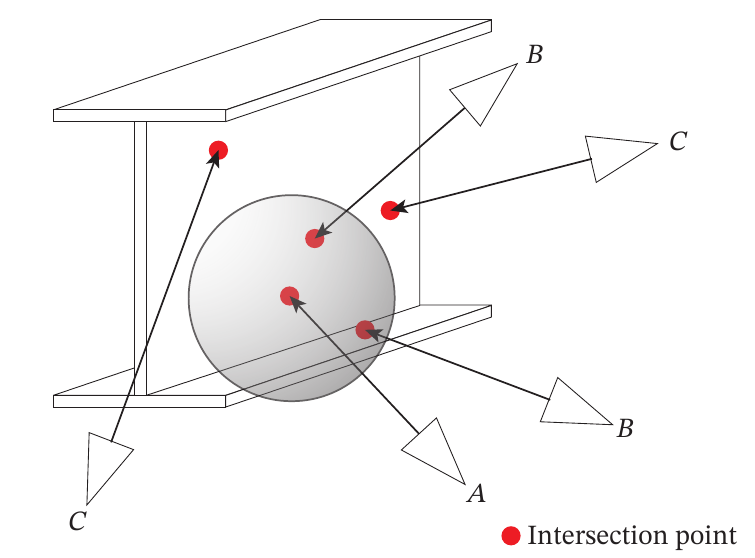}
\end{center}
   \caption{Image of interest and its surrounding images
   }
\label{fig:surroundingimg}
\end{figure}

To determine the surrounding images, we assume a ball centred on the intersection coordinates between the camera that captured the image of interest (Camera A) and the model, as shown in Fig.~\ref{fig:surroundingimg}. Cameras (Cameras B) whose intersection coordinates are inside the ball are treated as the images surrounding the image of interest. The other cameras (Cameras C) are considered to have little relevance to the image of interest and are not used in the subsequent analysis. In this study, for simplicity, the radius of this ball is assumed to be 1 m for determining the surrounding images. Although further discussion is needed on an appropriate setting for this value, the focus of this study is on the methodology itself, and we chose a value that is easy to calculate.

\section{Acquisition of damage information from bridge images}
\label{sec:acquisition of damage information}

\subsection{Visual Question Answering}
For damage cause estimation from images, it is necessary to obtain information on both damage and member as needed. VQA is a machine learning task that, given an image and a question related to the image, outputs an appropriate answer. The questions are input as sentences, and the answers are expected to be a wide range of content, such as words, yes/no, and numbers. The first dataset for the VQA task is the DAtaset for QUestion Answering on Real-world images (DAQUAR)~\cite{malinowski2014multi}, which is a dataset limited to indoor scenes with a total of 1449 images. Various other datasets have also been published, including Visual7W~\cite{zhu2016visual7w} and Visual Madlibs~\cite{yu2015visual}. In addition, datasets for VQA generally have multiple questions for a single image; Visual Question Answering (VQA)~\cite{antol2015vqa} is a widely used dataset for VQA, and an improved version of this dataset, Visual Question Answering v2.0 (VQA v2.0)~\cite{goyal2017making} with 265016 total images, is also available. Note that these datasets are constructed for general objects; therefore, it is difficult to utilize them directly in civil engineering fields such as bridge diagnostics.

As VQA is a task that has recently attracted attention with the development of deep leaning technology, deep-learning-based models have been mainly proposed. Most deep-learning-based models for VQA use CNNs~\cite{krizhevsky2017imagenet} for processing images, recurrent neural networks (RNNs)~\cite{rumelhart1986learning} for processing time series data, and Word2Vec~\cite{mikolov2013distributed} for word embeddings. In practice, not simple RNNs but their derivatives long short-term memory (LSTM)~\cite{hochreiter1997long} and gated recurrent units (GRUs)~\cite{cho2014learning} are often used.

There are several models that do not use RNNs. Zhou et al.~\cite{zhou2015simple} proposed a VQA model called iBOWIMG, which uses trained GoogLeNet~\cite{szegedy2015going} for image feature extraction and inputs questions as simple one-hot vectors to the model for feature extraction for each word. Ma et al.~\cite{ma2016learning} proposed a VQA model composed entirely of CNNs. Their proposed model consists of an Image CNN for processing images, a Sentence CNN for processing vectorized questions, and a multimodal convolution layer for combining the output from both CNNs.

Ren et al.~\cite{ren2015exploring} proposed a VQA model composed of a CNN and LSTM. In their model, the CNN-processed features on the image are first input to the LSTM, and then the words that make up the question are sequentially input to the LSTM. Malinowski et al.~\cite{malinowski2017ask} also proposed a VQA model consisting of a CNN and LSTM. Their model generates outputs by inputting CNN-processed features on images at the same time as words are sequentially inputted into the LSTM. Noh et al.~\cite{noh2016image} proposed DPPnet, a model consisting of a CNN and GRUs. This model outputs responses by integrating the obtained features by sequentially inputting each word into the GRUs and the features obtained from the image by the CNN into a dynamic parameter layer.

When using neural networks to handle sentences, simple RNN-based methods are limited in that accuracy deteriorates as the sentence length increases. Attention mechanism~\cite{bahdanau2014neural} has been proposed to focus on important parts by probabilistically weighting the input data. Attention mechanism has been extensively used in tasks such as sentence generation from images and machine translation. In recent years, attention mechanism has been widely used in image processing to improve accuracy by focusing on important parts of the input image. Many studies have been conducted in VQA to improve accuracy by focusing on important parts of input images and questions using the attention mechanism.

Shih et al.~\cite{shih2016look} presented a VQA model using attention mechanism. In their model, features from the input sentence are selectively combined with features from the image in a region selection layer to determine which parts of the image to focus on. Zhu et al.~\cite{zhu2016visual7w} proposed a VQA model using attention mechanism and LSTM. In their model, after a word is input to the LSTM, attention to the image is iteratively computed for each word. This attention region is determined by both the current word input to the model and the previous attention-weighted image. Lu et al.~\cite{lu2016hierarchical} proposed a co-attention model that performs attention not only to the image but also to the question text. In this model, an image representation is used to determine attention for the question text, and the question text representation is used to determine attention for the image.

As the co-attention model can consider the importance of each word compared to image-only attention, several co-attention-based VQA models have been proposed~\cite{nam2017dual,yu2018beyond}. However, as attention is determined independently for each modality (image and question), simple co-attention models do not consider the dense interaction between each word in the question and each image region. For example, attention to each word in a question is determined from the entire input image and may focus on areas that are not inherently important. Therefore, a dense co-attention model has been proposed to establish the full interaction between each word in a question and each image region~\cite{nguyen2018improved}. Compared to the co-attention model with coarse interaction, the dense co-attention model is more accurate. Another method that more accurately considers the interaction of each word with each image region is called bottom-up attention~\cite{anderson2018bottom}. Bottom-up attention uses a pre-trained object detection model to detect each object in the image and uses the results to calculate attention, which can consider the dense interaction of each word with each image region. Yu et al.~\cite{yu2019deep} proposed the bottom-up-attention-based Deep Modular Co-Attention Network (MCAN). MCAN consists of multiple modular co-attention (MCA) layers consisting of self- and guided-attention units. The self- and guided-attention units are constructed using the multi-head attention proposed by Vaswani et al.~\cite{vaswani2017attention}. Models with bottom-up attention are known to be significantly more accurate than non-bottom-up attention models.

\begin{figure}[h]
\begin{center}
\includegraphics[width=12cm]{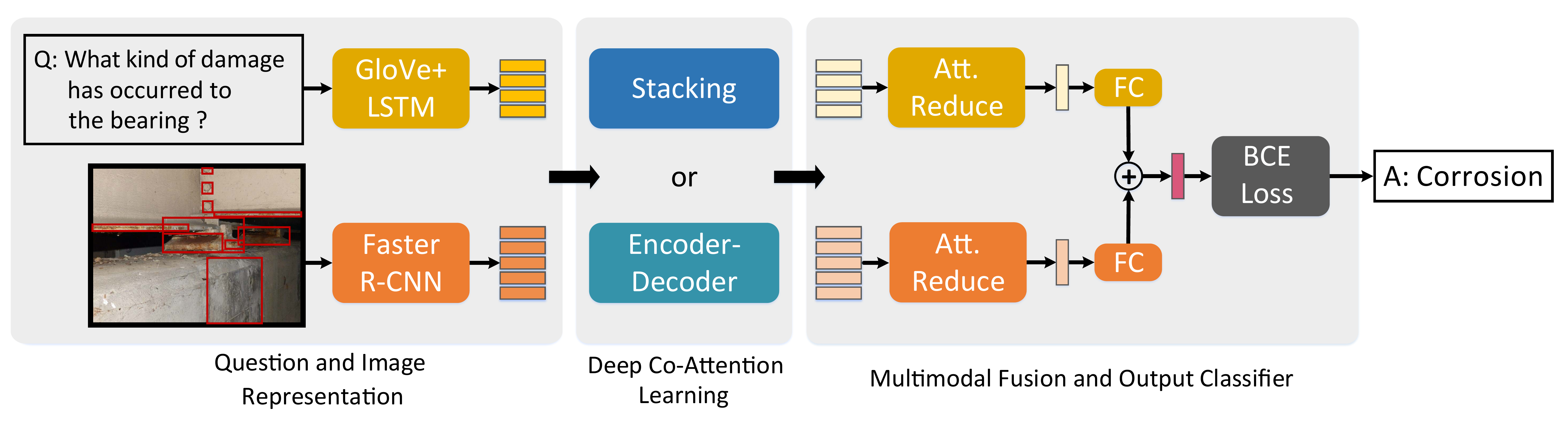}
\end{center}
   \caption{Deep modular co-attention network (MCAN) flowchart~\cite{yu2019deep}. 
   }
\label{fig:MCAN}
\end{figure}

For example, as shown in the MCAN flowchart in Fig.~\ref{fig:MCAN}, if the question is `What kind of damage has occurred to the bearing?' it is expected that the damage can be more accurately determined by focusing on the bearing, rather than on a location in the image that is unrelated to the content of the question. Therefore, in this study, a VQA model for bridge images was constructed using MCAN. It is worth noting that, as new VQA models are proposed, if a more accurate model is available, it can be easily adopted in the proposed framework.

\subsection{Training models using bridge images}
\subsubsection{Training dataset}

Various datasets for VQA are publicly available; however, these datasets are for general objects. Accordingly, even if a model is trained using such datasets, it is not possible to perform specialized VQA such as assessing bridge damage. Therefore, in this study, a new dataset for VQA was created using actual bridge images and a model was developed.

\begin{figure}[h]
\begin{center}
\includegraphics[width=12cm]{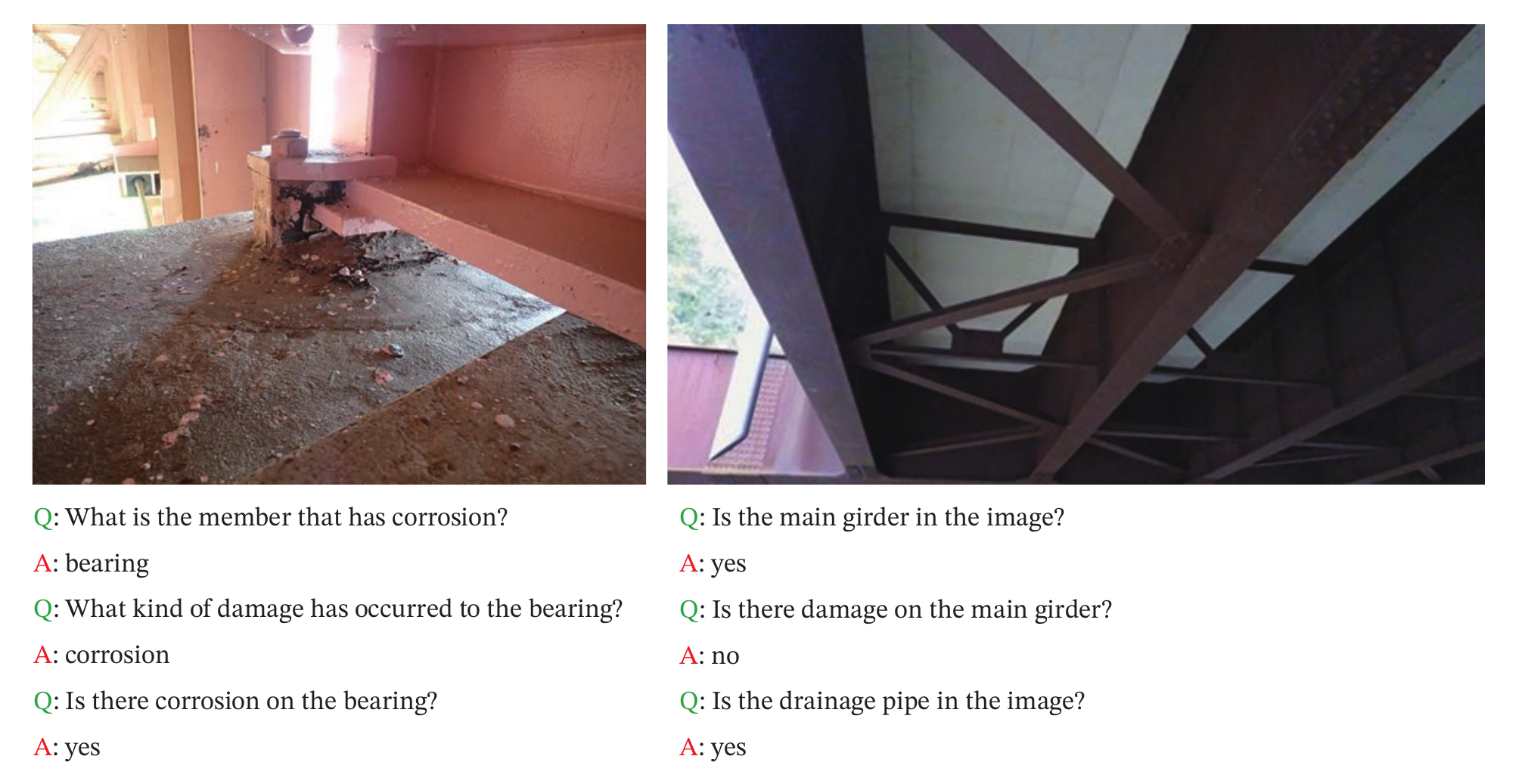}
\end{center}
   \caption{Example of the created dataset
   }
\label{fig:dataset}
\end{figure}

An example of the created dataset is shown in Fig.~\ref{fig:dataset}. For each image, questions were created asking what member is present, what damage is present, what damage occurs on what member, and so on.

Most of the images in the created dataset were extracted from actual bridge inspection reports at the Kanto Regional Development Bureau of the Ministry of Land, Infrastructure, Transport, and Tourism of Japan. However, bridge inspection reports are mainly composed of images taken of damaged areas; therefore, very few of these images do not show damage. By contrast, when unmanned aerial vehicles (UAVs) are used to capture images of actual bridges, it is expected that many undamaged images will be acquired. Therefore, in this study, in addition to the images extracted from bridge inspection reports, images extracted from videos of bridges taken by UAVs were added to the dataset. The number of images used to create the dataset was 442,035, consisting of 421,956 bridge inspection report images and 20,079 UAV-taken images.

When creating a VQA dataset, a human is usually free to create questions based on images, and the human creates appropriate answers based on the questions and images. However, when creating a VQA dataset from bridge images, specialized knowledge is required, and the cost of creating a VQA dataset is high if the same method for general datasets is used. Therefore, in this study, for images extracted from bridge inspection reports, questions and answers were mechanically generated based on the names of members and damages described in the inspection reports along with the images. For the UAV-extracted images, the names of the members in the images and the presence or absence of damage in the members were determined manually, and the questions and answers were generated mechanically based on the results.

\begin{figure}[t]
\begin{center}
\includegraphics[width=12cm]{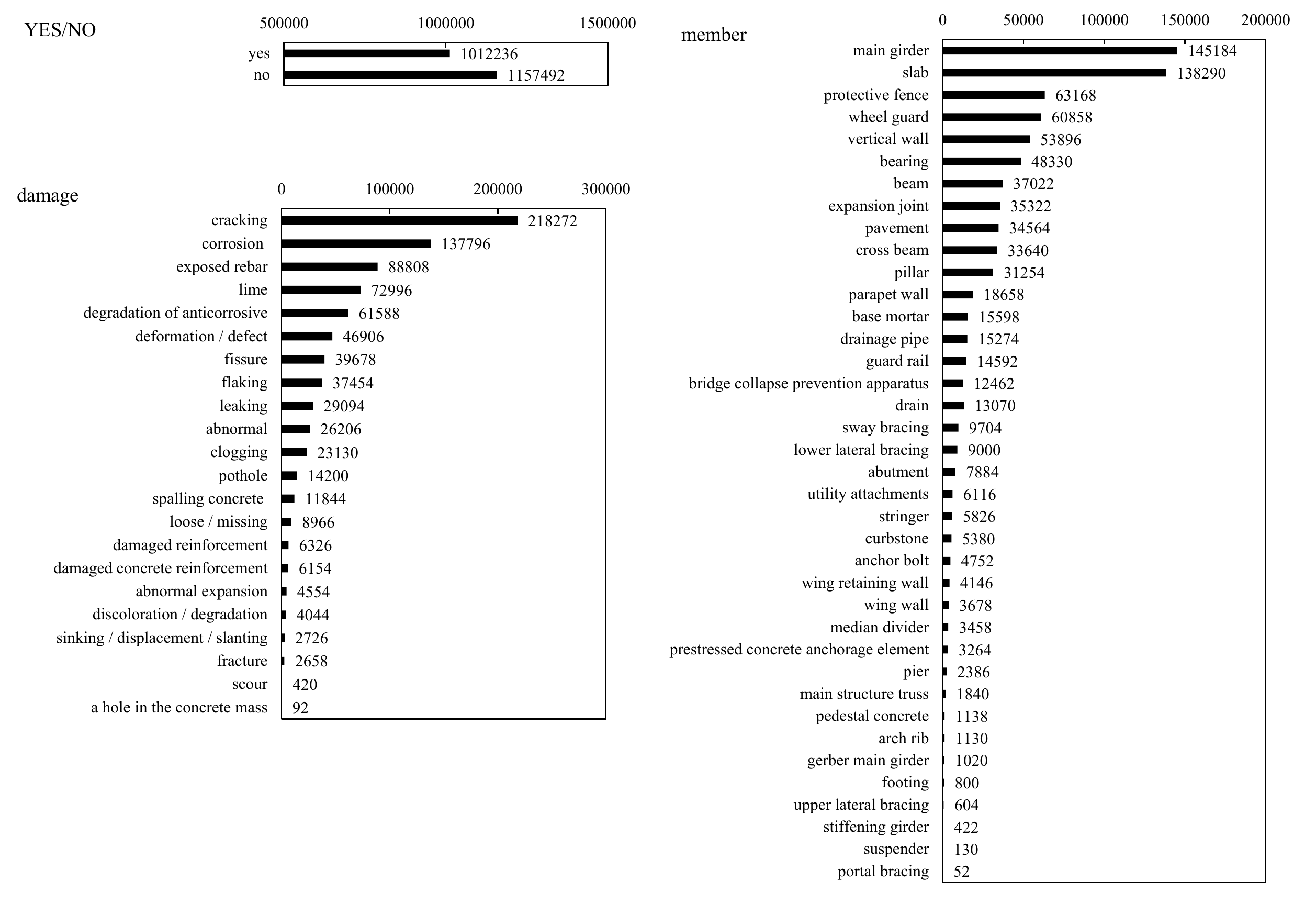}
\end{center}
   \caption{Breakdown of the answers in the dataset
   }
\label{fig:answer}
\end{figure}

Some damages in the inspection report were described as a single damage (e.g., sinking/displacement/slanting of bridge abutments). Therefore, such damages were treated as a single damage in the created dataset as well. In general VQA datasets, multiple questions are prepared for each image. Therefore, in this study, we created a dataset with multiple questions for each image. The total number of questions created was 3,875,010, with an average of approximately 8.8 questions per image. The dataset consists of questions that can be answered by name of the member and damage, and `yes/no' as shown in Fig.~\ref{fig:dataset}.  There are 38 types of members and 22 types of damages. Fig.~\ref{fig:answer} shows the occurrence of each answer in the dataset.

\subsubsection{Model training}

In this study, MCAN was trained on the created training data. The created dataset was divided into 310,259 images for the training data, 88,114 images for the validation data, and 43,662 images for the test data. Images of different bridges were used for each dataset. Adam~\cite{kingma2014adam} was used as the model optimization method; the Adam hyperparameters were $\beta_1$ = 0.9 and $\beta_2$ = 0.98, following Yu et al.~\cite{yu2019deep}. The learning rate was set to min(2.5$\textit{te}^{-5}$,1$\textit{e}^{-4}$), where \textit{t} is the current number of epochs starting from 1. This results in a small learning rate at the beginning of learning, which decays by 1/5 every two epochs after 10 epochs. The batch size at each epoch was 32, and training was performed up to 13 epochs.

In this study, the input images are entered as features obtained via bottom-up attention. These features are obtained as intermediate features extracted from a Faster R-CNN~\cite{ren2015faster} model (the backbone network is ResNet-101~\cite{he2016deep}) pre-trained on the Visual Genome dataset~\cite{krishna2017visual}. Note that the Visual Genome dataset does not include specialized data such as specific bridge member or damage names, but is a dataset for general objects. Therefore, when detecting objects in the image using the Faster R-CNN, the objects are detected as objects similar in shape to each member or damage.

\begin{figure}[h]
\begin{center}
\begin{minipage}[b]{0.45\linewidth}
\centering
\includegraphics[keepaspectratio, width=5cm]{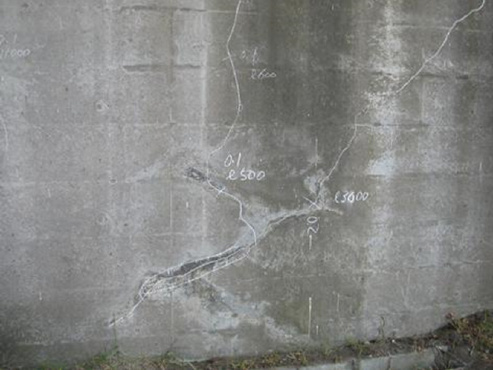}
\subcaption{Input image}
\end{minipage}
\begin{minipage}[b]{0.45\linewidth}
\centering
\includegraphics[keepaspectratio, width=5cm]{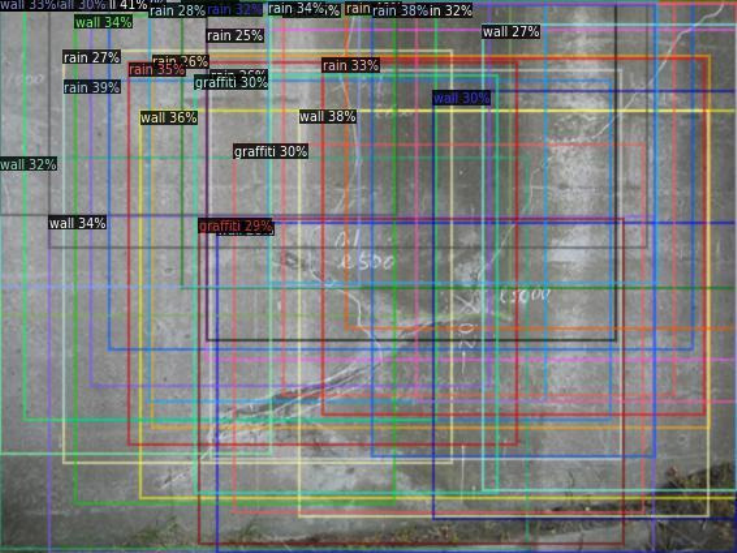}
\subcaption{Detection result}
\end{minipage}
\end{center}
\caption{Example of detection results using the Faster R-CNN
}
\label{fig:fasterrcnn}
\end{figure}

Fig.~\ref{fig:fasterrcnn} shows an object detection example from an image from the created dataset using the Faster R-CNN trained on the Visual Genome dataset. Fig.~\ref{fig:fasterrcnn} shows an image of a bridge abutment with water leaking, free lime, and cracks. The detection results show that `wall', `rain', and `graffiti' were detected. As shown in the figure, the Faster R-CNN trained on the Visual Genome dataset does not directly correctly recognize objects in its object detection task. However, similar detection results in our dataset indicate the presence of similar members or damage. For example, if another image with water leaking or free lime is input, the same `rain' or `graffiti' detection results are expected. Therefore, by using the features obtained here as input for MCAN, VQA based on the features in each image region can be performed without creating a separate object detection dataset for bridge inspection.

\subsubsection{Results and accuracy}

\begin{figure}[h]
\begin{center}
\includegraphics[width=12cm]{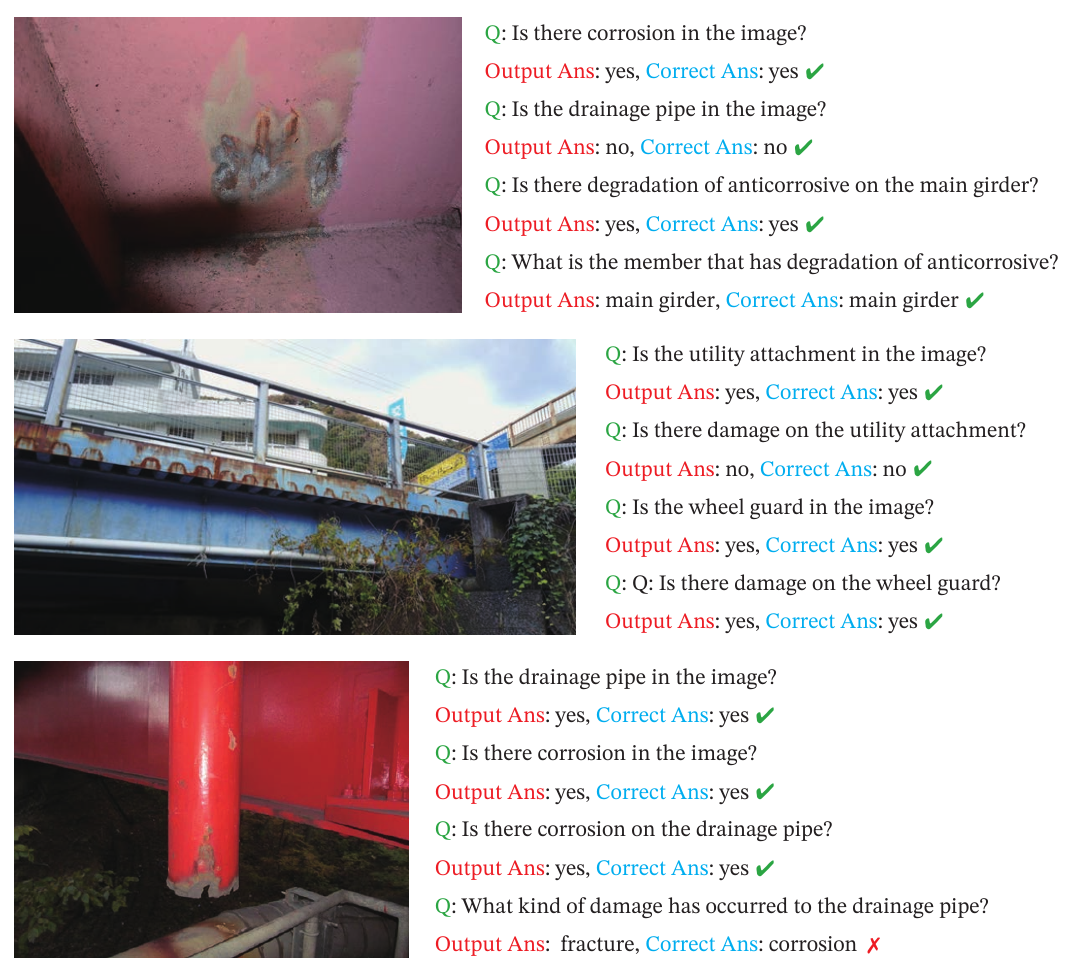}
\end{center}
   \caption{Example of output results from the trained model
   }
\label{fig:result}
\end{figure}

An example of the test data output results from the trained model is shown in Fig.~\ref{fig:result}. The top image in Fig.~\ref{fig:result} shows a main girder with corrosion and degradation of the anticorrosive material. In cases where this image and the question `Is there corrosion in the image?' were input into the model, the model correctly returned `yes'. In addition, the question `Is the drainage pipe in the image?' returned the correct result of `no'. Furthermore, the question `Is there degradation of the anticorrosive on the main girder?' also returned the correct result of `yes'. Finally, the question `What is the member that has degradation of the anticorrosive?' also returned the correct result, i.e., `main girder'.

The middle image in Fig.~\ref{fig:result} shows a girder bridge captured by a UAV. This bridge has utility attachments on the exterior girders, and the question `Is the utility attachment in the image?' was inputted, and `yes' was output correctly. In addition, the question `Is there damage on the utility attachment?' also returned the correct result of `no'. Furthermore, the question `Is the wheel guard in the image?' also returned the correct result, `yes'. Finally, the question `Is there damage on the wheel guard?' also returned the correct result, `yes'.

The lower image in Fig.~\ref{fig:result} shows a corroded drainage pipe. In cases where this image and the question `Is the drainage pipe in the image?' were input into the model, the model correctly returned `yes'. In addition, the question `Is there corrosion in the image?' also returned the correct result of `yes'. Furthermore, the question `Is there corrosion on the drainage pipe?' also returned the correct result, i.e., `yes'. However, the question `What kind of damage has occurred to the drainage pipe?' was returned as `fracture', which is incorrect. The drainage pipe in this image is not fractured and the tip is corroded, but fractured steel usually corrodes at the fracture point as well. Therefore, it is difficult to determine that the pipe has not fractured from this image alone. Nevertheless, the model answered most of the questions correctly.

The accuracy of the output results on the test data was evaluated using the trained VQA model. Table \ref{tab:correct} shows the percentage of correct answers for each output category on the test data. The correct answer rate for questions that require the member name as the answer was 67.4\%, and that for questions that require the damage name as the answer was 68.9\%. The question that required a yes/no answer had the highest percentage of correct answers, at 99.1\%. The overall correct answer rate for the test data was 81.2\%. 

\begin{table}[h]
    \caption{Percentage of correct output answers.
    }
    \begin{center}
        \begin{tabular}[!t]{ccccc}
            \toprule
            category & members & damages & YES/NO & All  \\
            \midrule
            Number of questions & 84392 & 84392 & 122926 & 291710 \\
            Number of correct answers & 56897 & 58152 & 121775 & 236824 \\
            Correct answer rate(\%) & 67.4 & 68.9 & 99.1 & 81.2 \\
            \bottomrule
        \end{tabular} 
    \end{center}
    \label{tab:correct}
\end{table}

One of the reasons for the higher accuracy of the yes/no questions is that the dataset created contains 38 types of members and 22 types of damages, resulting in more output options compared to yes/no questions. Furthermore, the low correct answer rate for questions that require the member name or damage as the answer is due to the fact that a single answer was created for each question; therefore, when multiple damages and members are included in a single image, they are considered to be incorrect.

\section{Damage cause estimation based on the acquired damage information}
\label{sec:estimation}

In this section, the damage cause estimation method based on image location calculated via SfM and the VQA results is described. In this study, the estimation method is proposed for corrosion as an example of damage for which the cause is to be estimated. When corrosion occurs in bridge members, the main cause is moisture supply from the surrounding area. There are many possible reasons and origins for this moisture. Typical reasons include the following:

\begin{itemize}
   \item Leaking from cracking on the slab;
   \item Leaking from cracking on the wheel guard;
   \item Leaking from damaged expansion joint;
   \item Leaking from damaged drainage pipe.
\end{itemize}

\begin{figure}[h]
\begin{center}
\includegraphics[width=12cm]{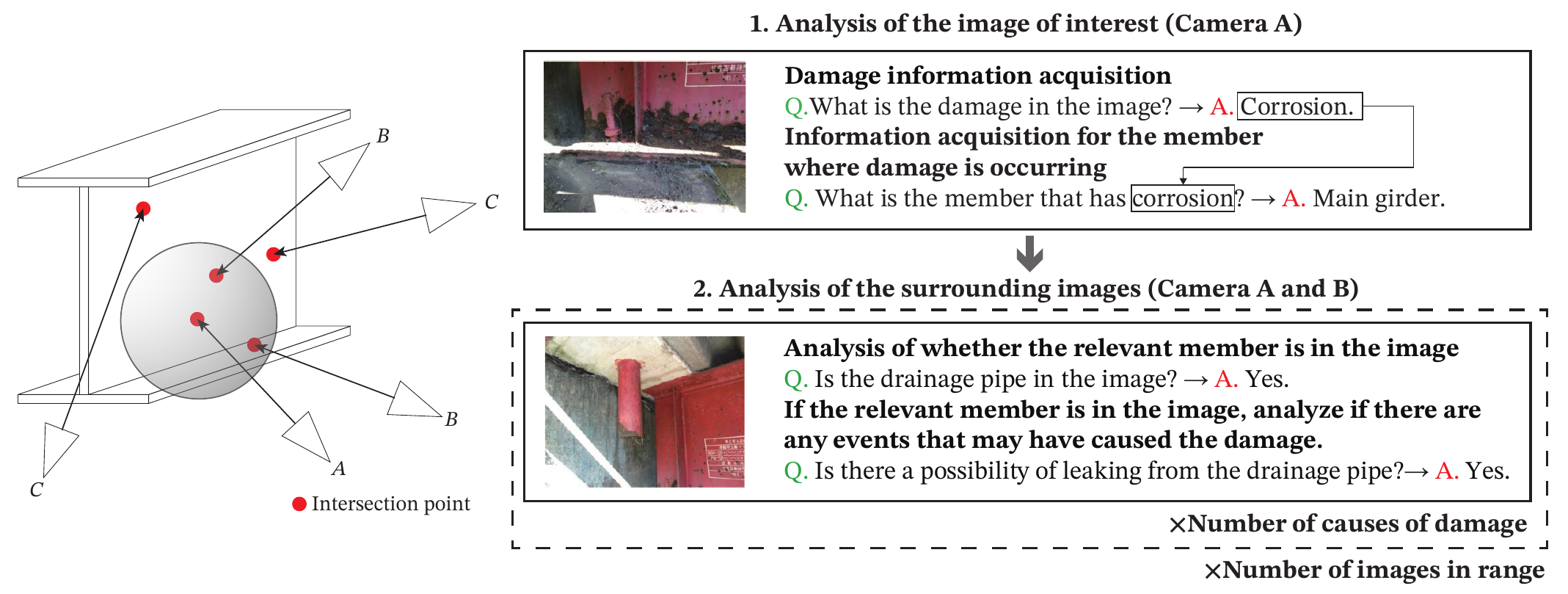}
\end{center}
   \caption{Damage cause estimation process
   }
\label{fig:estimation}
\end{figure}

To identify the cause of damage as described above, it is first necessary to determine whether or not these members exist in the vicinity of the corroded area. If a member is present, it is then necessary to check whether an event has occurred in that member that could have caused the damage. Based on this, the damage cause estimation process is as shown in Fig.~\ref{fig:estimation}.

As shown in Fig.~\ref{fig:estimation}, the first step is to analyse the image of interest (Camera A) (Step 1), where the type of damage in the image is determined. Then, based on the output damage name, a judgment is made regarding which member the damage is occurring in. Subsequently, based on the determined damage and member names, an analysis is performed to determine the damage cause that can be assumed in advance (Step 2). In this case, the surrounding images calculated via SfM are also used for analysis, as analysing the image of interest alone may miss information necessary for damage cause estimation. First, for the image to be analysed, it is determined if there are any members related to the assumed damage cause. If so, the system checks if there is an event that may have caused the damage to the member. In this way, the number of possible damage causes is analysed for each image. The above analysis is performed for all surrounding images, including the image of interest.

The number of images in which related members are determined to exist as a result of the analysis in Step 2 is \textit{N}. Let \textit{M} be the number of times that an event that could be the cause of damage is determined to occur. Then, the probability of each damage cause is expressed as \textit{M}/\textit{N}. If \textit{M}/\textit{N} is high, it is considered that the cause of damage is more likely to be the cause of the damage determined in Step 1.

In this way, damage causes can be estimated based on the SfM results and VQA. However, the cause of damage that can be estimated by this method is limited to cases where the damage is caused by the conditions surrounding the damage. Therefore, damage causes that are difficult to determine by analysing the surrounding images alone, such as salt damage, cannot be estimated. Such damage causes must be estimated in combination with other approaches, such as using a database of distances from shorelines.

\section{Field testing}
\label{sec:field test}
\begin{figure}[h]
\begin{center}
\includegraphics[width=6cm]{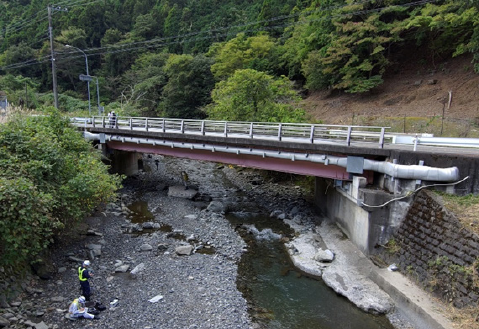}
\end{center}
   \caption{Steel girder bridge used for the field test in this study
   }
\label{fig:bridge}
\end{figure}

In this section, we verify whether the proposed framework can be used for damage cause estimation using actual bridge images. The bridge used in this study is the steel girder bridge shown in Fig.~\ref{fig:bridge}, which is managed by the Ministry of Land, Infrastructure, Transport, and Tourism of Japan. The span length and total width of the bridge are 25.4 and 9.8 m, respectively. The bridge was constructed in 1970 in the mountainous area of Yamanashi Prefecture, Japan. It has several damaged sections; particularly, there is severe corrosion on the superstructure.

\begin{figure}[h]
\begin{center}
\includegraphics[width=6cm]{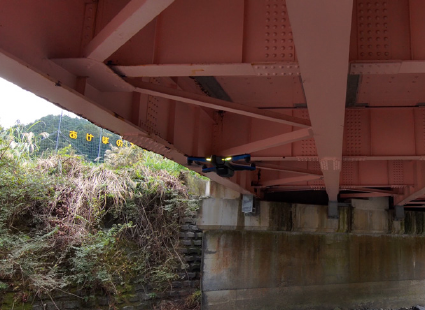}
\end{center}
   \caption{Bridge images taken using a small unmanned aerial vehicle
   }
\label{fig:uav}
\end{figure}

\begin{figure}[h]
\begin{center}
\includegraphics[width=6cm]{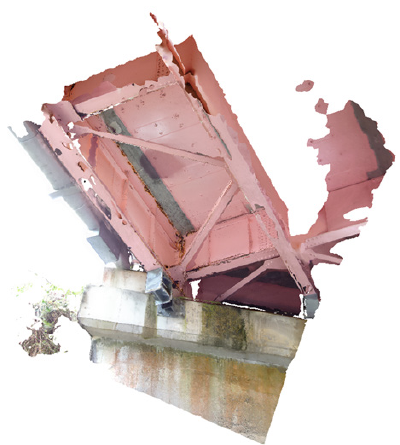}
\end{center}
   \caption{Polygon mesh model created based on the SfM results
   }
\label{fig:3dmodel}
\end{figure}

In this study, images of this bridge were taken using Skydio 2, a small UAV developed by Skydio, as shown in Fig.~\ref{fig:uav}. The focal length of the camera mounted on the UAV is approximately 4 mm, with a 35-mm-equivalent focal length of approximately 21 mm and a pixel count of approximately 12.3 megapixels (4056 × 3040 px). Then, Agisoft's Metashape was used for SfM processing. The number of images used for SfM was 541. The polygon mesh model created based on the SfM results is shown in Fig.~\ref{fig:3dmodel}.

\begin{figure}[h]
\begin{center}
\includegraphics[width=6cm]{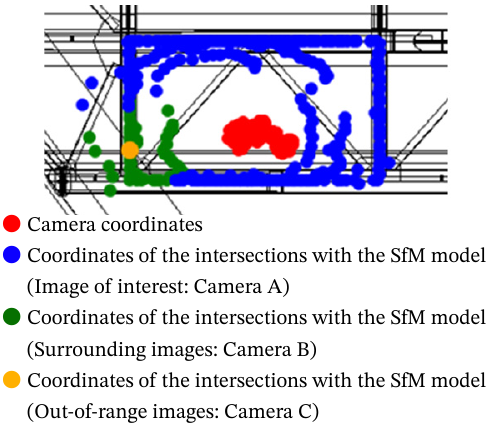}
\end{center}
   \caption{Camera and intersection coordinates
   }
\label{fig:coordinates}
\end{figure}

The coordinate positions of each camera and the intersection coordinates obtained from each camera are shown in Fig.~\ref{fig:coordinates}. The calculated coordinates are projected onto a two-dimensional plan view of the bridge. The red dots in Fig.~\ref{fig:coordinates} indicate the coordinate positions of each camera. The orange, green, and blue dots are the intersection coordinates calculated by intersection determination with the mesh model. The orange dots are the intersection coordinates of the image of interest used to estimate the cause of damage. The green points are the intersection coordinates that exist inside a sphere with a radius of 1 m centered on the intersection coordinates of the image of interest, and the number of images is 63. The blue dots are the intersection coordinates of the out-of-range images and are not used for estimating the cause of damage. In this study, the VQA model constructed in Sec.~\ref{sec:acquisition of damage information} was used to estimate the cause of damage based on the images indicated by the orange and green dots in Fig.~\ref{fig:coordinates}.

The coordinates of each camera and the intersection coordinates obtained from each camera are shown in Fig.~\ref{fig:coordinates}. The calculated coordinates are projected onto a two-dimensional plane view of the bridge. The red dots in Fig.~\ref{fig:coordinates} indicate the coordinates of each camera. The orange, green, and blue dots are the intersection coordinates calculated via intersection determination with the mesh model. The orange dots are the intersection coordinates of the image of interest used for damage cause estimation. The green dots are the intersection coordinates that exist inside a ball with a 1-m radius centred on the intersection coordinates of the image of interest (the number of images is 63). The blue dots are the intersection coordinates of the out-of-range images and are not used for damage cause estimation. In this study, the VQA model constructed in Sec.~\ref{sec:acquisition of damage information} was used for damage cause estimation based on the images indicated by the orange and green dots in Fig.~\ref{fig:coordinates}.

\begin{figure}[h]
\begin{center}
\includegraphics[width=6cm]{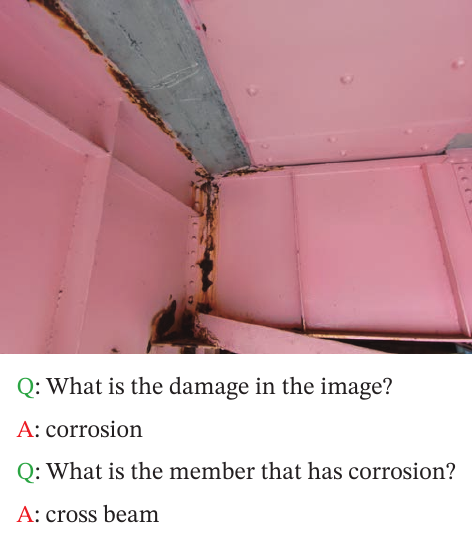}
\end{center}
   \caption{Output results for the image of interest
   }
\label{fig:result2}
\end{figure}

First, VQA was used to estimate the type of damage and the member in which it occurs in the image of interest. The analysis results are shown in Fig.~\ref{fig:result2}. First, a question was asked regarding what type of damage was occurring in the image of interest. In this case, the output result was `corrosion'. Based on the output results obtained, a question was then asked regarding which member was corroded. In this case, the result was a `cross beam'.

Then, VQA was used to analyse the image of interest and the surrounding images to estimate the cause of damage. In this study, four causes of corrosion damage were assumed and estimated: 1. leaking from cracking on the slab, 2. leaking from the expansion joint, 3. leaking from the drainage pipe, and 4. leaking from cracking on the wheel guard.

In determining water leaking through cracking on the slab, `slab' was set as a member related to the damage cause. The analysis was conducted assuming `cracking' and `leaking' as possible damage-causing events when a `slab' exists. In determining leaking from the expansion joint, `abutment' was set instead of expansion joint as the member related to the damage cause because the UAV could not approach the expansion joint in the range where the images were taken. This is valid because it is presumed that if there is leaking from the expansion joint, leaking can also be confirmed in the abutment directly below it. The analysis was conducted assuming `leaking' as a possible damage-causing event when an `abutment' exists. This allows the indirect determination of leaking from the expansion joint, despite the expansion joint not being photographed directly. In determining water leaking through the drainage pipe, `drainage pipe' was set as a member related to the damage cause. The analysis was conducted assuming `corrosion', `fissure', `fracture', and `leaking' as possible damage-causing events when a `drainage pipe' exists. Finally, in determining water leaking through cracking on the wheel guard, `wheel guard' was set as a member related to the damage cause. The analysis was conducted assuming `leaking' as a possible damage-causing event when a `wheel guard' exists.

\begin{figure}[h]
\begin{center}
\includegraphics[width=12cm]{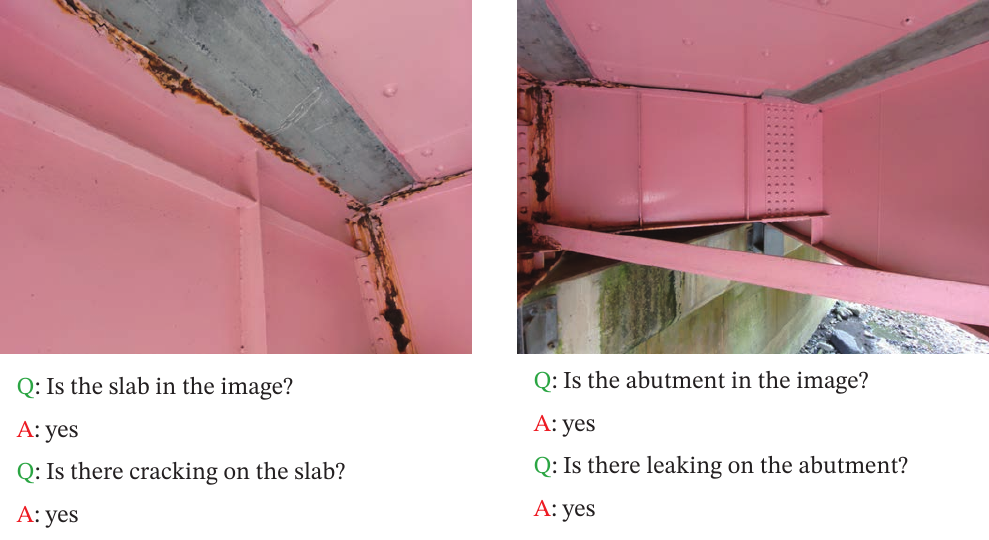}
\end{center}
   \caption{Example of output results with surrounding images
   }
\label{fig:result3}
\end{figure}

The analysis of these four damage causes was performed on 64 images, including the image of interest. An example of the analysis results is shown in Fig.~\ref{fig:result3}. The left image in Fig.~\ref{fig:result3} shows an example of checking for the presence of a slab in the image and then checking for cracking on the slab. The right image in Fig.~\ref{fig:result3} shows an example of checking for the presence of an abutment in the image and then checking for the presence of leaking marks on the abutment.

Table \ref{tab:results} shows the damage cause estimation results. The number of members \textit{N} in Table \ref{tab:results} indicates the number of images that were judged to contain related members that could cause corrosion. The number of damages \textit{M} indicates the number of images that are estimated to show damage that could cause corrosion among the images that are determined to show relevant members. As shown in Table \ref{tab:results}, out of the 64 images analysed, `slab' was determined to be present in 61 images, with `cracking' or `leaking' being detected on the slab in 58 of these images. `Abutment' was determined to be present in 22 images, with `leaking' being detected on the abutment in 18 of these images. The drainage pipe and wheel guard were not detected in any of the 64 images analysed in this study. Therefore, these were not considered to be the cause of damage in this case. In fact, the 64 images did not show any drainage pipes or wheel guards. Based on these results, corrosion is most likely caused by cracks on the slab or leakage from the expansion joint.

\begin{table}[h]
    \caption{Damage cause estimation results.
    }
    \begin{center}
        \begin{tabular}[!t]{ccccc}
            \toprule
            Damage & \multicolumn{4}{c}{Corrosion}  \\
            \midrule
            Member & \multicolumn{4}{c}{Cross beam}  \\
            \midrule
             & Leaking from & Leaking from the & Leaking from the & Leaking from\\
             Damage cause & cracking on & expansion joint	& drainage pipe & cracking on \\
             & the slab &&& the wheel guard \\
            \midrule
            \textit{N}: &&&& \\
            Number & 61 & 22 & 0 & 0 \\
            of members &&&& \\
            \textit{M}: &&&& \\
            Number & 58 & 18 & 0 & 0 \\
            of damages &&&& \\
            \textit{M}/\textit{N} & 0.95 & 0.82 & N/A & N/A \\
            \bottomrule
        \end{tabular} 
    \end{center}
    \label{tab:results}
\end{table}

Furthermore, it is confirmed that accurate damage cause estimation is possible from actual bridge images. It is worth noting that the image of interest shown in Fig.~\ref{fig:result2} does not show leaking on the abutment. Therefore, by using the proposed method to analyse multiple images for damage cause estimation, it is possible to obtain results that cannot be obtained by analysing the image of interest alone. Moreover, as the VQA model accuracy is not perfect, diagnosis based on a single image alone may yield erroneous results. However, as demonstrated here, diagnosis based on the analysis results of multiple images can provide a more accurate diagnosis. In addition, when practicing engineers use the proposed method, it is expected to contribute to preventing overlooking the cause of damage by checking the related members that are estimated to have a high damage cause probability by referring to the analysis results.

\section{Conclusion}
In this paper, a damage cause estimation method is proposed by calculating camera coordinates via SfM and extracting information from the images via VQA using multiple images taken around the damage region based on the calculated camera coordinates.

In damage cause estimation, it is desirable to utilize not only a single image but also images of the surrounding conditions. Therefore, we first described a method for acquiring the image of interest and its surrounding images using SfM. Next, a VQA model was constructed using MCAN with a dataset specific to bridge members and damage information. The correct answer rate for questions requiring the member's name and the damage's name were 67.4\% and 68.9\%, respectively. The correct answer rate for questions requiring a yes/no answer was 99.1\%. The overall correct answer rate for the test data was 81.2\%.

Furthermore, a damage cause estimation method is proposed using the SfM results and VQA model. Based on the VQA model output results and the pre-assumed damage causes, we proposed an approach to narrow the damage causes by inputting new questions into the VQA model.

The proposed method was used to determine the damage and estimate its cause on a real bridge. In this study, corrosion in bridge members was taken as the damage to be determined and estimate its causes. As a result, it was concluded that corrosion was most likely caused by cracks on the slab or leakage from the expansion joint.

When the proposed method is used by practicing engineers, it is expected to contribute to preventing overlooking the cause of damage by focusing on the related members that are estimated to have a high damage cause probability. In addition, for inexperienced engineers, it could also contribute to improve their maintenance skills, as they can learn where to focus in case of damage.

To improve the damage cause estimation accuracy, we could consider the severity of the damage, the relationship between the upper and lower positions of the damage, and weight the damage according to the distance between them. Further VQA model accuracy improvement is also important to improve the overall damage cause estimation accuracy.

\subsubsection{Acknowledgments}We would like to thank the Kofu River and National Highway Office, the Kanto Regional Development Bureau, and the Ministry of Land, Infrastructure, Transport, and Tourism of Japan for allowing us to take measurements of the bridges under their management. We would like to express our sincere gratitude to these organizations for making this research possible. The authors report there are no competing interests to declare. This research is supported by JST Moonshot Research and Development under Grant JPMJMS2032, and JSPS KAKENHI Grant Numbers 21H01417, 22J14364.

\bibliographystyle{splncs04}
\bibliography{egbib}

\end{document}